\documentclass[11pt,a4paper]{article}
\usepackage{cite} 
\usepackage[hyperref]{acl2021}
\usepackage{times}
\usepackage{latexsym}

\usepackage{bm}
\usepackage{amsmath,amssymb,mathrsfs}
\usepackage{graphicx}
\usepackage{CJKutf8}
\usepackage{multirow}
\usepackage{array}
\usepackage{tabularx}
\usepackage{makecell}
\usepackage{amsmath} 
\usepackage{amssymb}
\usepackage{tablefootnote}
\usepackage{subcaption}
\usepackage{diagbox}
\usepackage{algorithm}
\usepackage[noend]{algpseudocode}
\usepackage{booktabs}
\usepackage{xcolor}
\usepackage{multirow}
\usepackage{color, soul} %用color, 和 soul 包
\usepackage{colortbl}
\usepackage{makecell}
\usepackage{threeparttable} 

\usepackage{microtype}

\aclfinalcopy % Uncomment this line for the final submission
\newcolumntype{L}[1]{>{\raggedright\arraybackslash}p{#1}}
\newcolumntype{C}[1]{>{\centering\arraybackslash}p{#1}}
\newcolumntype{R}[1]{>{\raggedleft\arraybackslash}p{#1}}

\DeclareSymbolFont{extraup}{U}{zavm}{m}{n}
\DeclareMathSymbol{\varheart}{\mathalpha}{extraup}{86}
\DeclareMathSymbol{\vardiamond}{\mathalpha}{extraup}{87}

\title{Nested Named Entity Recognition as Holistic Structure Parsing}

\author{Yifei Yang$^{1,2,3}$, Zuchao Li$^{1,2,3}$, Hai Zhao$^{1,2,3,}$\thanks{$\ $  Corresponding author.}\\
$^{1}$Department of Computer Science and Engineering, Shanghai Jiao Tong University \\
	$^{2}$Key Laboratory of Shanghai Education Commission for Intelligent Interaction \\ and Cognitive Engineering, Shanghai Jiao Tong University, Shanghai, China\\
	$^{3}$MoE Key Lab of Artificial Intelligence, AI Institute, Shanghai Jiao Tong University \\
  {\tt \{yifeiyang,charlee\}@sjtu.edu.cn, zhaohai@cs.sjtu.edu.cn}}

\date{}

\begin{document}
\maketitle

\begin{abstract}
As a fundamental natural language processing task and one of core knowledge extraction techniques, named entity recognition (NER) is widely used to extract information from texts for downstream tasks. Nested NER is a branch of NER in which the named entities (NEs) are nested with each other. However, most of the previous studies on nested NER usually apply linear structure to model the nested NEs which are actually accommodated in a hierarchical structure. Thus in order to address this mismatch, this work models the full nested NEs in a sentence as a holistic structure, then we propose a holistic structure parsing algorithm to disclose the entire NEs once for all. Besides, there is no research on applying corpus-level information to NER currently. To make up for the loss of this information, we introduce Point-wise Mutual Information (PMI) and other frequency features from corpus-aware statistics for even better performance by holistic modeling from sentence-level to corpus-level. Experiments show that our model yields promising results on widely-used benchmarks which approach or even achieve state-of-the-art. Further empirical studies show that our proposed corpus-aware features can substantially improve NER domain adaptation, which demonstrates the surprising advantage of our proposed corpus-level holistic structure modeling.
\end{abstract}

\section{Introduction}
Named Entity Recognition is to find predefined named entities such as locations, organizations or people in text, which usually serves as an upstream natural language processing (NLP) task 
% (e.g., information extraction, event extraction, relation extraction, entity linking, sentiment analysis.) 
\citep{huang2015bidirectional,ma2016end,lample2016neural} and one of key knowledge extraction techniques in knowledge engineering.
% TODO

\begin{figure}[t]
    \centering
    \includegraphics[width=0.45\textwidth]{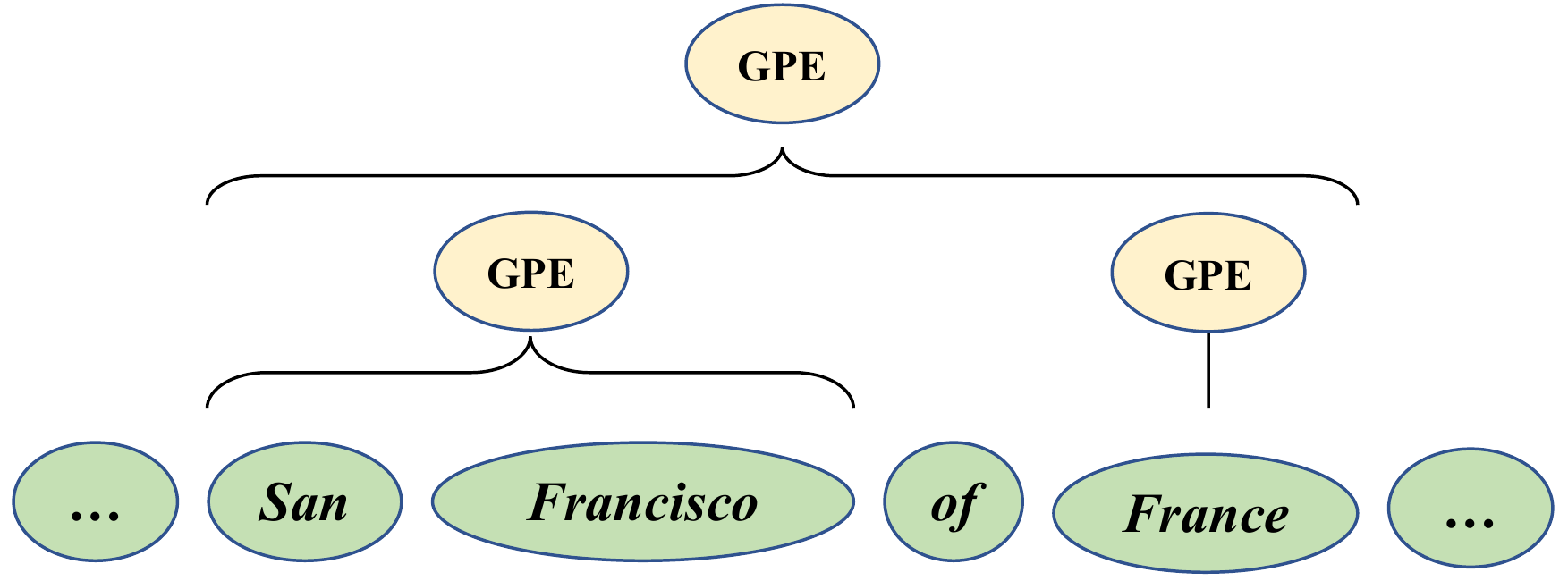}
    \caption{An example of nested named entities. The solid lines or brackets represents the start and end positions of a named entity. `GPE' is a named entity category, representing geographic or political entities. \emph{France} and \emph{San Francisco} are nested by \emph{the San Francisco of France} which forms a hierarchy structure.}
    \label{fig:case_study}
\end{figure}

Early works on NER \citep{ritter2011named,liu2011recognizing} mostly cope with only flat NEs that have no overlapped relationship with each other. However, as a common language phenomenon, nested NER appears universally in many corpora such as the field of biology or news events. An instance of overlapped NEs is shown in Figure\ref{fig:case_study}. Since solving the Nested NER task will bring more contextual information than the Flat NER task and promotes downstream tasks, it recently has aroused great research interest \citep{muis2018labeling,wang2018neural,luo2020bipartite}.

Although researchers have made good progress on flat NER, existing studies nested NER has not achieved a more satisfactory result yet.

\subsection{Limitations of Current Methods}
With the development of deep learning technology, nested NER has gradually developed from the methods based on handcraft features and traditional machine learning \citep{alex2007recognising,kumar2008hybrid} to the methods based on neural networks. In recent years, myriad studies apply neural network models to achieve state-of-the-art results.

From a perspective of deep learning, NER may be conveniently cast to a multi-class classification task or sequence labeling task \citep{fisher2019merge,lample2016neural,ma2016end}. Nevertheless, such modeling ways which mostly suit for flat NER cannot well handle nested NEs. Instead, there are two categories of models for nested NER, Layered-based Model and Region-based Model. The former recognizes the hierarchical structure by stacking multiple Flat NER layers. For examples, \citet{strakova2019neural} proposed a model with two stages, where the first stage is to discover the boundaries of the smallest granular NEs and the second one merges the smaller NEs into larger ones. \citet{wang-etal-2020-pyramid} proposed a model that consists of a stack of interconnected layers allowing higher layers to aggregate two adjacent hidden states from the lower layers. The latter type enumerates all text spans to discover the possible named entity mentions. For instances, \citet{lin2019sequence} predicted nested NEs by adopting Anchor Region Networks and \citet{sohrab2018deep} treated all adjacent tokens as potential NEs.
 
However, all of the above methods still view all nested NEs as a sequence structure and identify all the possible tags one by one for each NE span, which leads to two obvious drawbacks in modeling capability, (1) These methods require multiple modules to cooperate and each module has to be able to recognize NEs from a complete sentence, which usually results in a too complicated model and time-consuming processing \citep{luo2020bipartite,wang-etal-2020-pyramid}. (2) When relying on multiple modules working together, these methods further impose substantial information transmission among different modules, which inevitably causes serious error propagation \citep{ju2018neural}.

\subsection{Our Approach and Contributions}
% \subsection{Our Target Scenarios}
We address above limitation by modeling the nested NEs in a sentence as a holistic structure and propose a model which builds off work from recently constituency parsing algorithm \citep{kitaev2018constituency}. Thus, we extend our idea of holistic structure modeling from sentence-level to corpus-level by introducing corpus-aware features according to the statistics of point-wise mutual information or frequency over dataset and aggregating the information to our model by span attention \citep{tian2020improving}. The corpus-aware features hopefully enhances the current model by making up for such inability of the current representations.

Our experiments are conducted on three widely-used benchmark datasets following previous studies \citep{luo2020bipartite,wang-etal-2020-pyramid}. Results show that our model approaches or exceeds the current state-of-the-art with the newly-introduced methods.

Our contribution can be summarized into three-fold:
\begin{itemize}
\item We take a holistic structure to model all nested named entities in a sentence and adopt a chart parsing algorithm to decode every levels of nested named entities once for all, offering a concise and natural solution for so complicated a task.
\item We introduce corpus-aware features derived from PMI statistics to further boost model performance and enable the model a good domain adaptation capability.
\item With the simplicity of our modeling idea, our proposed holistic structure parsing model yields performance on par with or surpassing the state-of-the-art on widely-used English benchmark datasets.
\end{itemize}

\section{Model}
% This section presents our holistic structure modeling for nested NER and the PMI enhancement.
\subsection{Holistic Structure Modeling for Nested Named Entities}
\begin{figure}[t]
    \centering
    \includegraphics[width=0.43\textwidth]{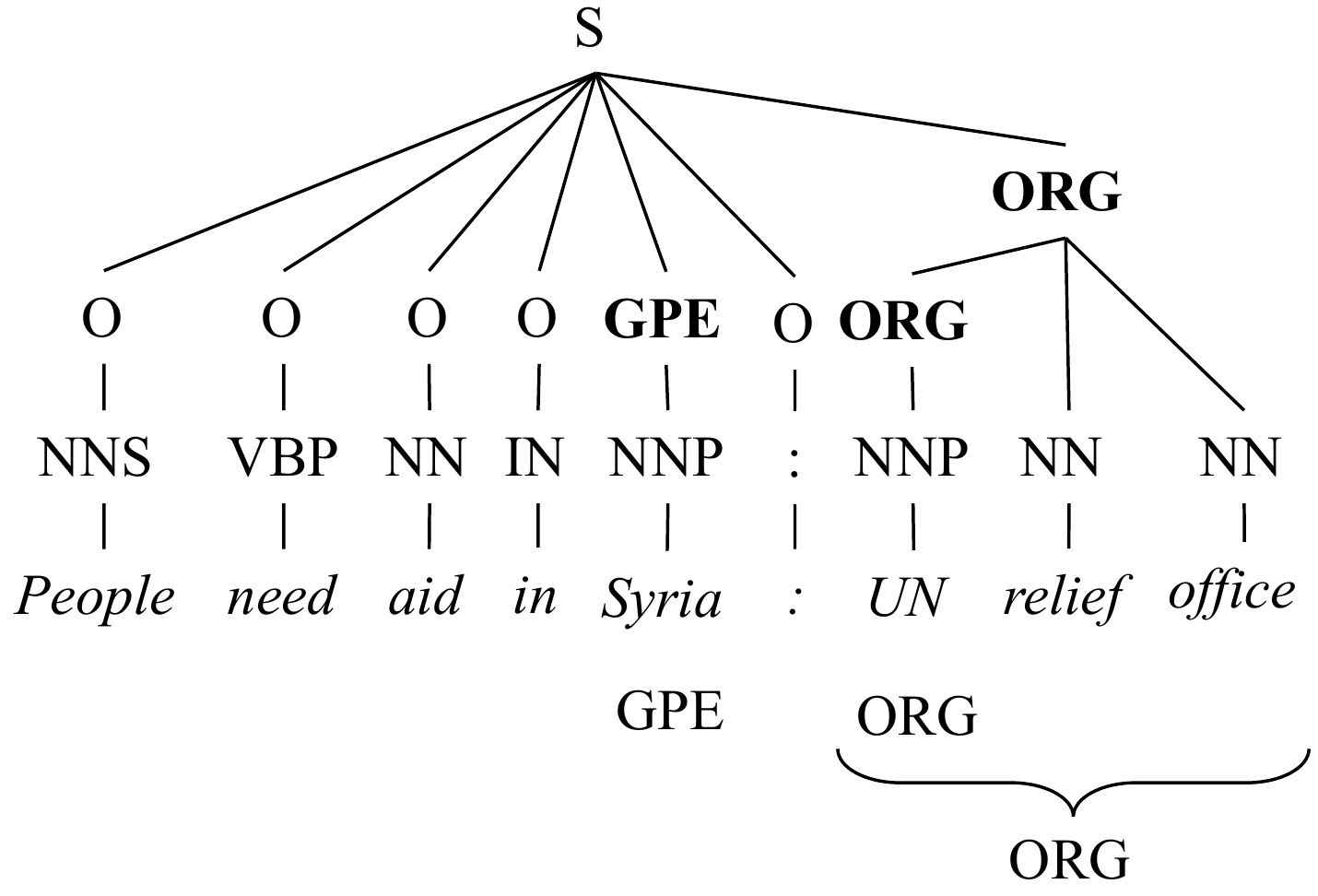}
    \caption{An example of nested NE tree. We use the tag `O' to label the span that is not recognized as a named entity. The nested named entities form is transformed by removing the tag `O'.}
    \label{fig:nested_NE_tree}
\end{figure}

Given a sentence of $n$ words $\mathcal{X}= \{x_1, x_2, ..., x_n\}$, we denote each named entity as a triplet $(i, j, l)$, where $i$, $j$ are the beginning and ending positions of a NE with label $l \in \mathcal{L}$, and $\mathcal{L}$ represents the set of NE types. 
Taking every levels of nested NE as constituent as syntactic parsing, all the labeled NE spans can form a tree structure like syntactic constituent parse tree (shown in Figure \ref{fig:nested_NE_tree}).

Our model assigns a real value score $s(i, j, l)$ to each labeled span. Then the score of a candidate tree $\mathcal{T}$ can be computed by the scores from all spans inside the tree:
\begin{align}
    s(\mathcal{T}) =\sum_{\begin{subarray}{c}(i,j,l) \in \mathcal{T}\end{subarray}} s(i,j,l).
\end{align}

Our model is trained and predicts the structure by selecting the tree $\widehat{\mathcal{T}}$ with the highest score by:
\begin{align}
    \widehat{\mathcal{T}} = \underset{\mathcal{T}}{\arg \max}\,s(\mathcal{T}).
\end{align}

Following \cite{gaddy-etal-2018-whats,taskar2005learning}, we apply margin-based training for such a structured prediction problem. Our model is trained to satisfy the constraint:
\begin{align}
    s(\mathcal{T^{*}}) \geq s(\mathcal{T}) + \Delta(\mathcal{T^{*}},\mathcal{T}),
\end{align}
\noindent where $\mathcal{T^{*}}$ is the golden tree and $\mathcal{T}$ covers all valid trees. $\Delta$ is the Hamming loss on labeled spans. Our objective function is defined by hinge loss:
\begin{equation}\nonumber
\max \left(0, \max _{\mathcal{T}}\left[s(\mathcal{T})+\Delta\left(\mathcal{T}, \mathcal{T^{*}}\right)\right]-s\left(\mathcal{T^{*}}\right)\right).
\end{equation}

% TODO 如果是用PPT画的可以考虑导出pdf，然后引用
\begin{figure*}[t]
    \centering
    \includegraphics[width=0.9\textwidth]{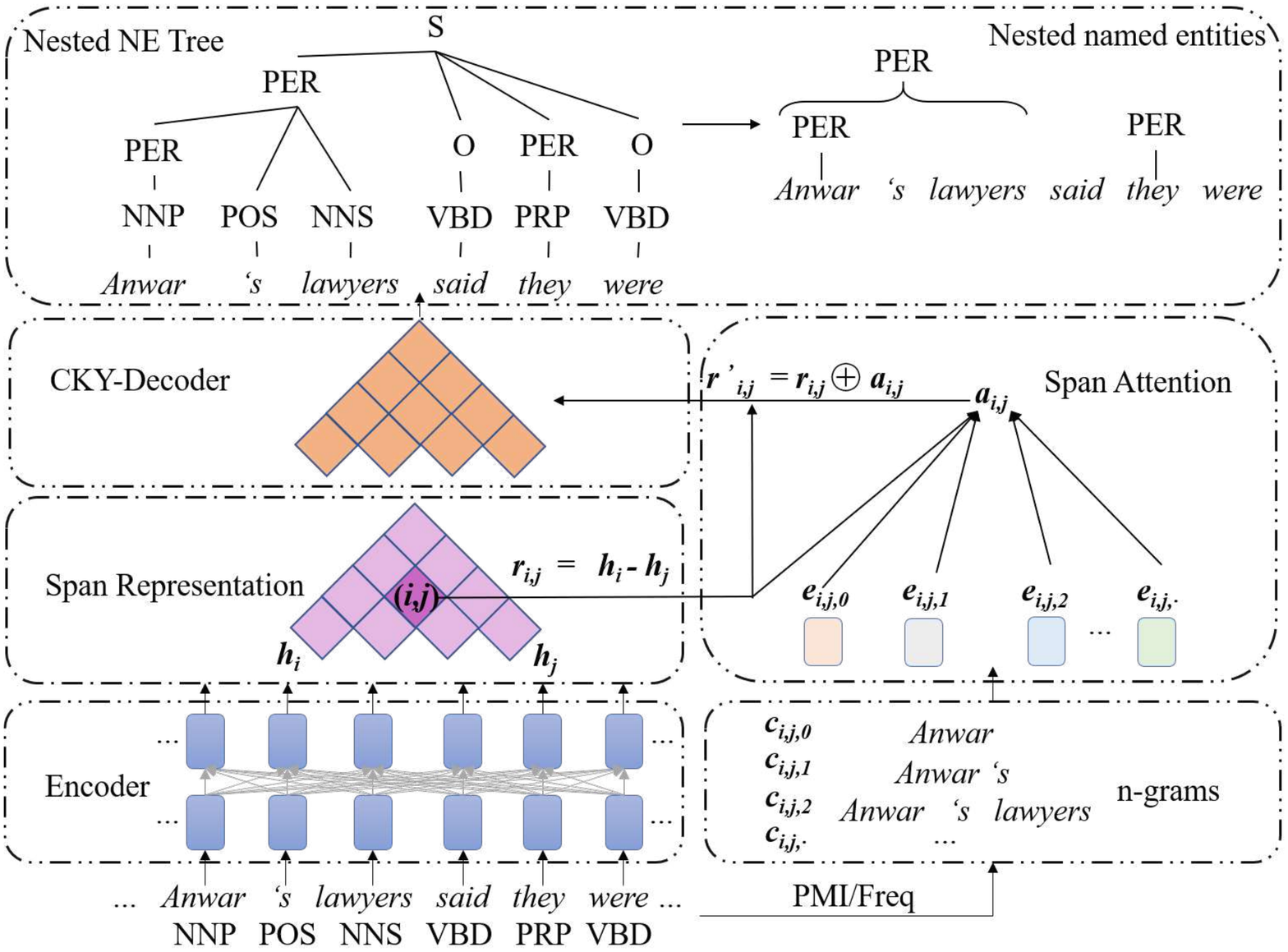}
    % \vspace{-5pt}
    \caption{The architecture of our model, where the encoder converts the input sentence into a set of hidden states and spans are represented based on them. Then the span representation and the span attention calculated by $n$-gram are concatenated together and entered into CKY-decoder.}
    \label{fig:model}
    \vskip -1em
\end{figure*}

\subsection{Contextualized Encoding}
Our model is implemented in an encoder-decoder framework.
For the encoder, it encodes each word $x_{t}$ in the input sentence $\mathcal{X}= \{x_1, x_2, ..., x_n\}$ into vectors $h_{t}$. Following \cite{fried2017improving,kitaev2018constituency}, $\mathcal{X}$ will be transformed into word embedding $\mathcal{W}= \{w_1, w_2, ..., w_n\}$ by embedding model. In addition, we embed part-of-speech (POS) tag and word position in sentence as $m_i$ and $p_i$ respectively for $x_i$. We set these three embeddings to the same dimension and then add them together to generate the input vector $z_{i}= w_{i}+m_{i}+p_{i}$ for the encoder.

We adopt the Transformer \cite{vaswani2017attention} as our encoder which is composed of 8 modules of the same self-attention structure. Each structure contains a multi-head attention $\textup{SubLayer}_{1}$ and a feed-forward $\textup{SubLayer}_{2}$. After each sublayer, there will be a residual connection and a Layer Normalization. So when given an input $x$, the output $y_{2}$ of each module is:
\begin{equation}
\setlength\abovedisplayskip{5pt}
\setlength\belowdisplayskip{5pt}
    y_{2} = \textup{LayerNorm}(y_{1} + \textup{SubLayer}_{2}(y_{1})),
\end{equation}
where $\textup{LayerNorm}$ denotes the Layer Normalization and $y_{1}$ is the output of the multi-head attention:
\begin{equation}
\setlength\abovedisplayskip{5pt}
\setlength\belowdisplayskip{5pt}
    y_{1} = \textup{LayerNorm}(x + \textup{SubLayer}_{1}(x)).
\end{equation}

After 8 stacked modules, the final output $h_t$ for word at position $t$ is generated.

\subsection{Decoder: Holistic Structure Parsing}
Our decoder performs the needed holistic structure parsing for every levels of nested NEs in a sentence.

After obtaining the context-aware representation $h_{t}$ for each word, the representation of NE span ($i$, $j$) is computed by:
\begin{equation}\label{eq: span_representation}
\setlength\abovedisplayskip{5pt}
\setlength\belowdisplayskip{5pt}
    r_{i,j} = h_{j} - h_{i},
\end{equation}
which assumes that the information of a subsequent step is generated by merging a previous state with the information of the span between them, so the difference between two steps can represent the span. In general, $s(i,j,\cdot)$ for span ($i$, $j$) over label set $\mathcal{L}$ is calculated by feeding $r_{i,j}$ into a multi-layer perceptrons (MLP), with ReLU activated function, which can be formalized by:
\begin{align} \label{eq: MLP_1}
\setlength\abovedisplayskip{5pt}
\setlength\belowdisplayskip{5pt}
        {o}_{i,j} &= \textup{ReLU}(\textup{LayerNorm}({W}_{1} \cdot {r}_{i,j} + {b}_{1})),
\end{align}
\noindent Finally, we have:
\begin{align} \label{eq: MLP_2}
\setlength\abovedisplayskip{5pt}
\setlength\belowdisplayskip{5pt}
    s(i,j, \cdot) &= {W}_{2} \cdot {o}_{i,j} + {b}_{2},
\end{align}
where ${W}_{1}$, ${W}_{2}$ and ${b}_{1}$, ${b}_{2}$ are all trainable parameters.

We adopt CKY parsing algorithm \cite{younger1967recognition} as our decoding algorithm to search for the highest score $s(i,j)$ for span ($i$, $j$). In details, to begin with the span ($i,i+1$) of length 1, we only need to consider its label:
\begin{equation} \label{eq: best (i,j)}
\setlength\abovedisplayskip{5pt}
\setlength\belowdisplayskip{5pt}
    \begin{array}{ll}
        s^{*}(i,i+1) = \max\limits_{l \in \mathcal{L}} s(i,i+1,l).
    \end{array}
\end{equation}

Then, we can extract the highest score $s^{*}(i,j)$ for longer span ($i$, $j$) in a recursive way by searching for the best matching label $l$ and the best boundary $k$:
\begin{equation} \label{eq: best (i,j)}
\setlength\abovedisplayskip{5pt}
\setlength\belowdisplayskip{5pt}
    \begin{array}{ll}
        s^{*}(i,j) &= \max\limits_{l \in \mathcal{L}} s(i,j,l) \\
        &+ \max\limits_{i < k < j}[s^{*}(i,k) + s^{*}(k,j)].
    \end{array}
\end{equation}

In this way, for the current span, the best label and best split point are chosen separately to find the highest score. 

According to the above steps, for an entire sentence, we use a bottom-up approach to find the highest score $s^{*}(0,n)$ which is the sum of the highest scores for its subtrees. The full structure for the entire sentence can be rehabilitated by traversing backpointers. Thus we parse the sentence as a holistic structure and disclose all nested named entities by giving the nested NE tree once for all.

As the same as classical chart parsing, the computational complexity for the CKY algorithm to parse a sentence of length $n$ is $O(n^3)$, which is better than enumerating all possible spans in $O(n^4)$.

\section{Corpus-aware Features}

Though our proposed holistic structure parsing model at sentence-level hopefully captures the entire sentence for predicting all nested NEs, which is supposed to yield more powerful model capability than previous models, it still has the limitation similar with the previous ones which only learn the representation inside a sentence. Meanwhile, NEs quite sparsely distribute in all text data such as most NEs only occur for once in the entire corpus. Besides, there may be a huge domain difference among various corpora, which lets the model recognize new NEs even more difficultly. Thus, we further introduce corpus-aware clues to alleviate the above mentioned difficulties.

In detail, we consider corpus-level statistics including frequency and point-wise mutual information (PMI) over words and corresponding POS tags. Therefore, we consider the following statistics to form enhanced features, PMI on word (Word PMI), word frequency (Word Freq), PMI on POS tags (POS PMI) and on POS tags frequency (POS Freq).

Equipped with the corpus-aware features from a large amount of unlabeled corpus, the model may be enabled to effectively adapt to different domains by alleviating the difficulty caused by the sparsity of NEs distribution.

\subsection{Point-wise Mutual Information and Frequency}
The PMI between two adjacent tokens $x'$,$x''$ is calculated as:
\begin{equation}
\mathbf{PMI}(x', x'') = \log \frac{p(x'x'')}{p(x')p(x'')} 
\end{equation}

We calculate the PMI between two adjacent words in turn and split a sentence from the low PMI positions to obtain multiple $n$-grams.

Similar with the above Word PMI processing, we can also count every $n$-grams ($n$\textless 10) to record their frequency in the corpus, then perform sentence segmentation according to a pre-specified threshold $t_{freq}$ following the same processing to obtain the results of Word Freq.

In view of the sparsity of $n$-gram occurrence, $n$-gram statistics like the above Word PMI or Word Freq may still encounter the issue of data sparsity. Thus we turn to the more informative POS which usually has a limited sized POS tag set \cite{diab2007improved,benajiba2007anersys}.

To apply the POS information, we first perform POS tagging over every sentences in the corpus and save the correspondence between the word and the POS tag. Then we segment the spans of POS tags according to PMI or frequency. Finally, we convert these spans back to their corresponding word $n$-grams.

\subsection{Integration of Corpus-aware Features}\label{app: Integration_feature}
Following \cite{tian2020improving}, we incorporate corpus-aware features into our model through span attention scoring to update the original NE span score in Eq. (\ref{eq: span_representation}).

Before training our model, we extract all the $n$-grams from the training set and development set by Word PMI, Word Freq, POS PMI, POS Freq and store them in a Lexicon $\mathcal{N}$. Given a sentence $\mathcal{X}$, we enumerate all the spans ($i$, $j$) in it and find the spans which are included by $\mathcal{N}$ to generate a set $\mathcal{C}_{i,j} = \{c_{i,j,1}, c_{i,j,2}, \cdots c_{i,j,v}, \cdots c_{i,j, m_{i,j}}\}$.
By span attention, each $n$-gram $c_{i,j,v}$ in $\mathcal{C}_{i,j}$ will be given an attention score by:

\begin{equation} \label{eq: a_ijv}
\setlength\abovedisplayskip{5pt}
\setlength\belowdisplayskip{5pt}
    a_{i,j,v} = \frac{\exp(\mathbf{r}_{i,j}^{\top} \cdot \mathbf{e}_{i,j,v})}
                    {\sum_{v=1}^{m_{i,j}} \exp(\mathbf{r}_{i,j}^{\top} \cdot \mathbf{e}_{i,j,v})},
\end{equation}
where $\mathbf{e}_{i,j,v} \in \mathbb{R}^{d_{r}}$ refers to the embedding of $c_{i,j,v}$.
By the weighted average of $n$-gram embeddings, the resulted attention of span ($i$, $j$) is:

\begin{equation}
\label{eq: a_ij}
\setlength\abovedisplayskip{5pt}
    \mathbf{a}_{i,j} = \sum_{v=1}^{m_{i,j}} a_{i,j,v} \mathbf{e}_{i,j,v}.
\end{equation}

As there are more short $n$-grams in the corpus than long $n$-grams, it is necessary to encourage these $n$-grams of various lengths in a balanced way, since there is no difference in the probability of being an entity.
Thus, we split $n$-grams by their lengths into different categories, i.e., $\mathcal{C}_{i,j}=\{\mathcal{C}_{i,j,1}, \mathcal{C}_{i,j,2}, \cdots \mathcal{C}_{i,j,u}, \cdots \mathcal{C}_{i,j,n}\}$, where $u \in [1, n]$ indicates the $n$-gram length and weight them by each category following Eq. (\ref{eq: a_ijv}-\ref{eq: a_ij}) to calculate $a^{(u)}_{i,j,v}$ and $\mathbf{a}^{(u)}_{i,j}$. 
Thus the final attention of the split $n$-grams is composed of the cascade all category attentions:

\begin{equation}\label{eq: a_iju}
    \mathbf{a}_{i,j} = \mathop{\oplus}_{1\leq u\leq n} \delta_{u} \mathbf{a}^{(u)}_{i,j}, 
\end{equation}
where the weight of attentions from different categories is counterpoised by trainable parameter $\delta_{u}$.
After $\mathbf{a}_{i,j}$ is calculated for span ($i$, $j$), our model concatenates it with $r_{i,j}$ (generated by Eq. \ref{eq: span_representation}): $\mathbf{r}'_{i,j}=\mathbf{r}_{i,j} \oplus \mathbf{a}_{i,j} \in \mathbb{R}^{d_{r} \cdot (n+1)}$. Then $r'_{i,j}$ will be used to get $s(i,j,\cdot)$ by Eq. (\ref{eq: MLP_1}-\ref{eq: MLP_2}).

\begin{table*}[!htb]
\setlength\tabcolsep{9pt}
    \centering
    \resizebox{0.99\linewidth}{!}{
    \begin{tabular}{lccccccccc}
    \toprule
    \multirow{2}{*}{\bf Model}  & \multicolumn{3}{c}{{\bf ACE 2005}} & \multicolumn{3}{c}{{\bf GENIA}} & \multicolumn{3}{c}{{\bf KBP 2017}}\\
    \cmidrule(lr){2-4} \cmidrule(lr){5-7} \cmidrule(lr){8-10}& P & R & F$_{1}$ & P & R & F$_{1}$ & P & R & F$_{1}$\\
    \midrule 
    Hyper-Graph \citep{katiyar2018nested} & 70.60 & 70.40 & 70.50 & 77.70 & 71.80 & 74.60 & $-$ & $-$ & $-$ \\
    Seg-Graph \citep{wang2018neural} & 76.80 & 72.30 & 74.50 & $-$ & $-$ & $-$ & $-$ & $-$ & $-$ \\
    ARN \citep{lin2019sequence} & 76.20 & 73.60 & 74.90 & 75.80 & 73.90 & 74.80 & 77.70 & 71.80 & 74.60 \\
    Merge-BERT \citep{fisher2019merge} & 82.70 & 82.10 & 82.40 & $-$ & $-$ & $-$ & $-$ & $-$ & $-$\\
    DYGIE \citep{luan2019general} & $-$ & $-$ & 82.90 & $-$ & $-$ & 76.20 & $-$ & $-$ & $-$ \\
    Seq2seq-BERT \citep{strakova2019neural} & $-$ & $-$ &  84.33 & $-$ & $-$ &  78.31 & $-$ & $-$ & $-$\\ 
    Path-BERT \citep{shibuya2019nested} & 82.98 & 82.42 & 82.70 & 78.07 & 76.45 & 77.25 & $-$ & $-$ & $-$ \\ 
    BERT-MRC \citep{li2019unified} & 87.16 & 86.59 & 86.88 & {\bf 85.18} & {\bf 81.12} & {\bf 83.75} & 82.33 & 77.61 & 80.97 \\
    Seq2seq-BART \citep{yan2021unified} & 83.16 & 86.38 & 84.74 & 78.57 & 79.30 & 78.93 & $-$ & $-$ &  $-$ \\
    Seq2set-BERT \citep{tansequence} & {\bf 87.48} & 86.63 & 87.00 & 82.31 & 78.66 & 80.40 & 84.91 & 83.04 &  83.90 \\

    \midrule
    \bf Ours & 86.81 & {\bf 88.70} & {\bf 87.75} & 79.76 & 75.74 & 77.70 & {\bf 88.25} & {\bf 87.10} & {\bf 87.67}\\
    \bf \quad-Corpus-Feat. & 84.65 & 87.19 & 85.90 & 79.74 & 74.13 & 76.83 & 86.62 & 86.61 & 86.62\\
    \bottomrule
    \end{tabular}}
    \caption{Results for nested NER tasks.}
    \label{main results}
\end{table*}

\section{Experiments}

\subsection{Setup}
We evaluate our approach on three nested named entity recognition benchmark datasets: GENIA, ACE2005 and KBP2017 datasets.

% The data statistics are shown in Table \ref{tab:data_stat}.

\textbf{ACE2005} \cite{walker2006ace} contains 25\% nested named entities and 7 entity types. We apply the same setup as \cite{lu2015joint,katiyar2018nested,wang2018neural} by splitting the dataset into training/development/test sets by 8:1:1, respectively.

\textbf{GENIA} dataset \cite{kim2003genia} is based on the GENIAcorpus3.02p\footnote{http://www.geniaproject.org/genia-corpus/posannotation}. The dataset contains 10\% nested mentions and 5 entity types. We follow the same train/dev/test as previous work \cite{finkel2009nested,lu2015joint} and split first 81\% as training set, subsequent 9\% as development set, and last 10\% as test set.

\textbf{KBP2017} contains 19\% nested entities. We evaluate our model on the Event Nugget Detection Evaluation dataset (LDC2017E55) following \cite{lin2019sequence}, and previous annotated datasets (LDC2015E29, LDC2015E68, LDC2016E31 and LDC2017E02) are added into  training and development sets. We split the datasets into 866/20/167 documents for training, development and test.

We use Precision (P), recall (R) and F-score (F$_{1}$) to evaluate the predicted named entities and a predicted named entity is regarded as correct if it exists in the golden labels.

\begin{table*}[!htp]
\centering
\small
\resizebox{0.99\linewidth}{!}{
\begin{tabular}{ccccccccccccccccc}
    \toprule
    \multirow{2}{*}{\bf DATA}  & \multirow{2}{*}{\bf Pre-trained}  & \multicolumn{3}{c}{\textbf{Word PMI}} & \multicolumn{3}{c}{\textbf{Word Freq.}} & \multicolumn{3}{c}{\textbf{POS PMI}} & \multicolumn{3}{c}{\textbf{POS Freq.}} & 
    \multicolumn{3}{c}{\textbf{-Corpus-Feat.}} \\ 
    \cmidrule(lr){3-5}\cmidrule(lr){6-8} \cmidrule(lr){9-11}\cmidrule(lr){12-14} \cmidrule(lr){15-17}& & P & R  & F$_1$ & P   & R & F$_1$ & P & R  & F$_1$ & P   & R & F$_1$ & P   & R & F$_1$ \\ 
    % \midrule
    % \multirow{4}{*}{CoNLL03}& GloVe & \\
    % & BERT$_{base}$ & 90.48 & 90.18 & \cellcolor[HTML]{EFEFEF}90.34 & 90.37 & 90.61 & \cellcolor[HTML]{EFEFEF}90.49 \\ 
    % & BERT$_{wwm}$ & \\
    % & XLNet$_{large}$ & \\
    \midrule
    \multirow{4}{*}{ACE2005} & Random Init. & 70.10 & 48.68 & \cellcolor[HTML]{EFEFEF} 57.46$^{*}$ & 73.35 & 46.10 & \cellcolor[HTML]{EFEFEF}56.62 & 72.13 & 49.31 & \cellcolor[HTML]{EFEFEF} {\bf 58.58$^{\dagger}$} & 74.33 & 48.22 & \cellcolor[HTML]{EFEFEF}56.53 & 68.35 & 47.21 & \cellcolor[HTML]{EFEFEF}55.85 \\
    & BERT$_{base}$ & 82.27 & 83.15 & \cellcolor[HTML]{EFEFEF} {\bf 82.71$^{*}$} & 80.94 & 83.05 & \cellcolor[HTML]{EFEFEF}81.98 & 81.47 & 82.61 & \cellcolor[HTML]{EFEFEF} 82.04$^{\dagger}$ & 81.25 & 82.31 & \cellcolor[HTML]{EFEFEF}81.78 & 78.06 & 79.34 & \cellcolor[HTML]{EFEFEF}78.69\\ 
    & BERT$_{wwm}$ & 84.05 & 85.47 & \cellcolor[HTML]{EFEFEF} {\bf 84.75$^{*}$} & 83.42 & 85.29 & \cellcolor[HTML]{EFEFEF}84.34 & 83.42 & 85.29 & \cellcolor[HTML]{EFEFEF} 84.34$^{\dagger}$ & 82.75 & 84.22 & \cellcolor[HTML]{EFEFEF}83.48 & 82.38 & 83.93 & \cellcolor[HTML]{EFEFEF}83.15\\
    & XLNet$_{large}$ & 86.81 & 88.70 & \cellcolor[HTML]{EFEFEF} {\bf 87.75$^{*}$} & 84.89 & 87.16 & \cellcolor[HTML]{EFEFEF}86.01 & 85.23 & 87.19 & \cellcolor[HTML]{EFEFEF} 86.20$^{\dagger}$ & 84.95 & 87.40 & \cellcolor[HTML]{EFEFEF}86.16 & 84.65 & 87.19 & \cellcolor[HTML]{EFEFEF}85.90 \\
    \midrule
    \multirow{4}{*}{GENIA}  & Random Init. & 64.21 & 31.46 & \cellcolor[HTML]{EFEFEF} 42.23$^{*}$ & 71.05 & 29.44 & \cellcolor[HTML]{EFEFEF}41.63 & 61.85 & 43.97 & \cellcolor[HTML]{EFEFEF}51.40 & 73.03 & 47.28 & \cellcolor[HTML]{EFEFEF} {\bf 57.40$^{\dagger}$} & 60.05 & 28.29 & \cellcolor[HTML]{EFEFEF}38.46 \\
    & BERT$_{base}$ & 78.71 & 73.44 & \cellcolor[HTML]{EFEFEF} {\bf 75.99$^{*}$} & 77.60 & 72.43 & \cellcolor[HTML]{EFEFEF}74.92 & 77.75 & 73.57 & \cellcolor[HTML]{EFEFEF} 75.60$^{\dagger}$ & 77.67 & 71.60 & \cellcolor[HTML]{EFEFEF}74.51 & 76.86 & 69.38 & \cellcolor[HTML]{EFEFEF}72.93 \\
    & BERT$_{wwm}$ & 79.07 & 74.92 & \cellcolor[HTML]{EFEFEF} {\bf 76.94$^{*}$} & 78.79 & 74.26 & \cellcolor[HTML]{EFEFEF}76.46 & 78.38 & 74.76 & \cellcolor[HTML]{EFEFEF} 76.53$^{\dagger}$ & 77.05 & 74.56 & \cellcolor[HTML]{EFEFEF}75.79  & 76.11 & 74.01 & \cellcolor[HTML]{EFEFEF}75.05\\
    & XLNet$_{large}$ & 79.84 & 75.43 & \cellcolor[HTML]{EFEFEF}77.57 & 79.82 & 75.66 & \cellcolor[HTML]{EFEFEF} 77.69$^{*}$ & 79.76 & 75.74 & \cellcolor[HTML]{EFEFEF} {\bf 77.70$^{\dagger}$} & 79.21 & 75.30 & \cellcolor[HTML]{EFEFEF}77.21 & 79.74 & 74.13 & \cellcolor[HTML]{EFEFEF}76.83\\
    \midrule
    \multirow{4}{*}{KBP2017} & Random Init. & 75.73 & 49.32 &  \cellcolor[HTML]{EFEFEF} 59.73$^{*}$ & 76.14 & 40.76 &  \cellcolor[HTML]{EFEFEF}53.10 & 72.41 & 45.56 & \cellcolor[HTML]{EFEFEF}55.93 & 77.74 & 50.54 & \cellcolor[HTML]{EFEFEF} {\bf 61.26$^{\dagger}$} & 70.21 & 40.32 & \cellcolor[HTML]{EFEFEF}51.22 \\
    & BERT$_{base}$ & 83.45 & 80.22 & \cellcolor[HTML]{EFEFEF}81.80 & 82.95 & 81.18 & \cellcolor[HTML]{EFEFEF} {\bf 82.06$^{*}$} & 82.42 & 80.10 & \cellcolor[HTML]{EFEFEF} 81.24$^{\dagger}$ & 82.46 & 78.76 & \cellcolor[HTML]{EFEFEF}80.57 & 76.34 & 74.66 & \cellcolor[HTML]{EFEFEF}75.49 \\
    & BERT$_{wwm}$ & 85.04 & 83.31 & \cellcolor[HTML]{EFEFEF}84.17 & 85.80 & 84.00 & \cellcolor[HTML]{EFEFEF} {\bf 84.89$^{*}$} & 85.56 & 82.83 & \cellcolor[HTML]{EFEFEF} 84.17$^{\dagger}$ & 82.91 & 79.35 & \cellcolor[HTML]{EFEFEF}81.09 & 80.45 & 79.48 & \cellcolor[HTML]{EFEFEF}79.96\\
    & XLNet$_{large}$ & 88.15 & 86.48 & \cellcolor[HTML]{EFEFEF} 87.30$^{*}$ & 87.96 & 86.41 & \cellcolor[HTML]{EFEFEF}87.18 & 88.25 & 87.10 & \cellcolor[HTML]{EFEFEF} {\bf 87.67$^{\dagger}$} & 87.68 & 86.32 & \cellcolor[HTML]{EFEFEF}87.00 & 86.62 & 86.61 & \cellcolor[HTML]{EFEFEF}86.62\\ 
    \bottomrule
\end{tabular}}
\caption{Ablation study on the multiple corpus-aware features inducing approaches. $^{*}$ and $^{\dagger}$ indicate the higher F$_{1}$ in Word PMI, Word Freq and POS PMI, POS Freq. Bold numbers indicate the highest F$_{1}$ in an experiment.}
\label{tab:unsup_feat_abalation}
\end{table*}

\subsection{Implementation}
In our experiments, we use randomly initialized embedding, variants of BERT \cite{devlin2018bert} and XLNet$_{large}$ \cite{yang-2019-xlnet} to embed sentences. 
Following \cite{tian2020improving}, we initialize all $n$-gram embeddings in the span attention module and match their dimensions with the hidden vectors from the encoder. 
During the training process, we use Adam optimizer with the learning rate 5e-5, 1e-5, and 5e-6. We also fine-tune dropout to attention, POS tag embedding and residual with the rate of 0.2, 0.4, 0.5. We select the model with the highest F$_{1}$ on the development set and evaluate it on the test set.

We set $t_{pmi}$ to 0 to determine which words to form $n$-grams and 0.5 for POS tags since some POS tags often appear next to each other, but their corresponding words are meaningless. Such as the POS tag pair $(, \textnormal{VBD})$ has a high frequency of co-occurrence but the corresponding span $(, \textnormal{had})$ is pointless, thus we set a higher threshold to filter this situation. Similarly, we set $t_{freq}$ to 2 for words and 5 for POS tags.

\subsection{Main Results}

Table \ref{main results} compares our results with other state-of-the-art approaches on three nested NER benchmark datasets. By adding corpus-aware feature, our model has a significant improvement on KBP2017 which exceeds the state-of-the-art model \citep{tansequence} by 3.77 F$_{1}$, 3.34 P and 4.06 R. We also receive a remarkable advancement on ACE2005 and outperform the state-of-the-art by 0.75 F$_{1}$ and 2.07 R. On the GENIA dataset, even though our model performs relatively unsatisfactorily, it still outperforms better than most models shown in Table \ref{main results}. 

After removing the corpus-aware features of each model while the other training parameters remaining unchanged, the results show that the F$_{1}$ of all models receive a drop greatly, which proves that our corpus-aware features indeed enhance the effect. Although without them, our method also surpasses the previous best results on the KBP2017 by a large margin and surmounts most preceding approaches on ACE2005 and GENIA, which confirms modeling Nested NER as holistic structure is essentially feasible.

\section{Ablation Study}

\subsection{Effect of different corpus-aware features}

For exploring the effects of the four corpus-aware features with significance test, we carry out 12 sets of experiments. For each set of them, we combine different features and pre-trained language models to conduct comparative analysis.

The results are shown in Table \ref{tab:unsup_feat_abalation}. For each dataset, we use a randomly initialized word embedding and three pre-training language models to experiment separately and each pre-training language model has been trained with four corpus-aware features. During the training process, we allow the model to fine-tune the pre-trained language model in order to get better result. In order to better verify the effect of them, we also conduct a set of controlled experiment ablating corpus-aware features as a comparison.

In the comparison between PMI and Freq, the PMI-based features devote 8 highest F$_{1}$ in all 12 experiments while the Freq-based features occupy the remaining 4 highest results. Namely, PMI-based features show more helpful than Freq-based features for both Word and POS types, which implies PMI is indeed a better statistical measure than the common Freq for building informative corpus-aware features. The result also shows that text-level co-occurrence knowledge has more information than vanilla frequency.

\subsection{Role of POS Tags}

\begin{table}[!htbp]
\setlength\tabcolsep{9pt}%调列距
    \centering
    \tiny
    \resizebox{0.99\linewidth}{!}{
    \begin{tabular}{lccc}
         \toprule
         \bf Method & P & R & F$_1$ \\
         \midrule
         \bf Ours & 88.25 & 87.10 & 87.67\\
         \quad-POS embedding & 87.64 & 86.21 & 86.92\\
         \quad-POS PMI & 86.62 & 86.61 & 86.62\\
         \quad-Both & 86.31 & 85.89 & 86.10\\
         \bottomrule
    \end{tabular}}
    \caption{Ablation study on POS tags.}
    \label{tab:pos_abalation}
\end{table}

For the comparison of Word and POS, the Word-based features attain 7 highest F$_{1}$ while the POS-based features gain the rest 5 highest results. This phenomenon displays that Word-based features are slightly better than POS-based features. This conclusion is also in line with common perception that words themselves provide more knowledge than their POS tags.

The performance obtained by removing corpus-aware features is always the calamity which likewise supports the significant improvement of our proposed corpus-aware features. From overall view, the Word PMI has the best lifting effect for our model, followed by POS PMI. Word Freq and POS Freq are slightly less effective while both of them share the similar effect. Meanwhile, the proposed corpus-aware features can be easily imposed into any models that require span extraction.

\subsection{Domain Adaptation Effects}
\begin{table*}[htp]
\setlength\tabcolsep{9pt}%调列距
    \centering
    \small
    \resizebox{0.99\linewidth}{!}{
    \begin{tabular}{llllllllll}
    \toprule
    
    \multirow{2}{*}{\bf Model}  & \multicolumn{3}{c}{{\bf ACE 2005}} & \multicolumn{3}{c}{{\bf GENIA}} & \multicolumn{3}{c}{{\bf KBP 2017}}\\
    \cmidrule(lr){2-4} \cmidrule(lr){5-7} \cmidrule(lr){8-10}& {P} & {R} & {F$_{1}$} & {P} & {R} & {F$_{1}$} & {P} & {R} & {F$_{1}$}\\
    \midrule
    ACE-trained Model & 86.81 & 88.70 & 87.75 & 24.28$^{*}$ & 10.19$^{*}$ & 14.36$^{*}$ & 73.89 & 76.06 & 74.96\\
    GENIA-trained Model & 6.49$^{*}$ & 0.10$^{*}$ & 0.19$^{*}$ & 79.76 & 75.74 & 77.70 & 51.66$^{*}$ & 2.18$^{*}$ & 4.18$^{*}$\\
    KBP-trained Model & 73.49 & 71.78 & 72.62 & 5.88$^{*}$ & 0.88$^{*}$ & 1.54$^{*}$ & 88.25 & 87.10 & 87.67\\
    \midrule
    \multicolumn{10}{c}{\textit{ACE-trained Model}} \\
    + GENIA Unlabeled Data & 86.51 & 87.96 & 87.23 & 27.36$^{*}$ & 14.09$^{*}$ & {\bf 18.60$^{*}$} & 74.40 & 74.60 & 74.50 \\
    + KBP Unlabeled Data & 85.72 & 87.50 & 86.60 & 22.50$^{*}$ & 7.91$^{*}$ & 11.71$^{*}$ & 74.66 & 75.40 & {\bf 75.03$^{*}$}\\
    \midrule
    \multicolumn{10}{c}{\textit{GENIA-trained Model}} \\
    + ACE Unlabeled Data & 50.00$^{*}$ & 1.20$^{*}$ & {\bf 2.35$^{*}$} & 79.38 & 74.80 & 77.02 & 37.64$^{*}$ & 0.81$^{*}$ & 1.59$^{*}$\\
    + KBP Unlabeled Data & 51.05$^{*}$ & 2.53$^{*}$ & 4.82$^{*}$ & 78.13 & 74.01 & 76.01 & 52.02$^{*}$ & 3.13$^{*}$ & {\bf 5.90$^{*}$}\\
    \midrule
    \multicolumn{10}{c}{\textit{KBP-trained Model}} \\
    + ACE Unlabeled Data & 73.87 & 71.85 & {\bf 72.85} & 16.60$^{*}$ & 3.86$^{*}$ & 6.26$^{*}$ & 87.28 & 86.86 & 87.07\\
    + GENIA Unlabeled Data & 74.19 & 70.91 & 72.51 & 17.01$^{*}$ & 4.28$^{*}$ & {\bf 6.83$^{*}$} & 86.46 & 84.79 & 85.61 \\
    \bottomrule
    \end{tabular}}
    \caption{Results for domain adaptation experiments. $^*$ indicates unlabeled F$_{1}$ due to different named entity annotation standards. Bold numbers represent an improvement compared to the baselines.}
    \label{cross nested NER results}
\end{table*}

In order to verify that the corpus-aware features can enhance the domain adaptability of our model, we conduct cross-domain experiments between the three datasets. We find the best trained model on one dataset and evaluate the effect of this model on other datasets as our baseline (shown at the top in Table \ref{cross nested NER results}). Subsequently, we retrain the best trained model without changing any hyper-parameter setting and add external unlabeled data separately. In details, the model extracts $n$-gram from the external unlabeled data by leveraging the corpus-aware features and blends them into the lexicon $N$ gleaned from the labeled training dataset. The changed lexicon $N$ will be used to generate span attention and the domain adaptation ability of our model will be affected subsequently.

The lower part of Table \ref{cross nested NER results} shows the results. It can be inferred that when a best trained model is joined a non-NER labeled dataset, the prediction effect of the model on this dataset will improve. The biggest improvement occurs when the KBP-trained and ACE-trained model are added GENIA non-NER labeled data. We deduce that the huge increments are because ACE and KBP both derive from the event domain while GENIA comes from a different biological domain. The model will perform poorly in a domain which is pretty different from the training corpus. Once the knowledge of that domain is compensated through the corpus-aware feature, the performance of the model in that domain will be greatly improved. This result verifies the effectiveness of our corpus-aware features for better domain adaptation.

We unexpectedly discover that the domain adaptation between ACE and KBP datasets has not made effective progress because they share the same domain. The corpus-aware features can not bring additional useful knowledge when the external unlabeled datasets are in the same domain with the training corpora. This inference is in line with our intuition and likewise demonstrates that our proposed corpus-aware features are able to extract information from different domain. We also found a counterintuitive fact which is that GENIA-trained model with KBP unlabeled data outperforms the GENIA-trained model with ACE unlabeled data on ACE. The explanation we give is that the knowledge in the news domain brought by KBP unlabeled data is more suitable for this GENIA trained model, which makes the model achieves a greater improvement on the ACE dataset.

% 同时我们也发现了一个反常counterintuitive.的现象，GENIA-trained Model with KBP Unlabeled Data outperforms the GENIA-trained Model with ACE Unlabeled Data on ACE 2005, 我们认为这仍然是因为这两个数据集的domain是近似的，因此unlabeled corpus feature也是相似的，而这组实验中引入的KBP Unlabeled data更适合于提升GENIA-trained Model在新闻领域数据集上识别的准确度。

In summary, the corpus-aware features can not only enhance the effect of the model but also can effectively improve its domain adaptation ability.

% \subsection{Effects of Different Unlabeled Data Scales}
% 跑出点画一张图

\section{Related Work}

The current Nested NER approaches can be classified into three classes.

\begin{itemize}
    \item \textbf{Hypergraph model.} This approach is a common method applied for the nested named entities whose main idea is to apply hypergraph structure to represent nested NEs. \citet{lu2015joint} first proposed a hypergraph-based method by connecting multiple nodes with edges to represent nested NEs. \citet{muis2018labeling} further employed multigraph representation and introduced a novel notion of mention separators to detect nested named entity.
    However, these methods are not intelligent enough due to the need to manually design explicit hypergraph.
    
    \item \textbf{Stacking layered model.} This is also a widespread approach to handle nested NER. \citet{alex2007recognising} stacked several neural layers to recognize the lower and higher level NEs separately. \citet{fisher2019merge} proposed a neural network that merges NEs or tokens to generate nested structure and labels them. Several recent researches applied multi-layer GCN to accomplish nested NER \citep{li2021span,luo2020bipartite}. These are practical methods, nonetheless, the existing models with a huge depth are computationally impractical.
    
    \item \textbf{Region-based model.} This method is another common method for nested NER, which enumerates all the possible subsequences and recognizes them. \citet{sohrab2018deep} regarded all spans as potential NEs and used neural networks to recognize them. A layered model that enumerates all the potential subsequences while preserving the sequence structure was presented by \citep{wang-etal-2020-pyramid}. The biggest disadvantage of this method is that the model has computational complexity of at least $O(n^4)$ for the sentence of length $n$. Thus, this method requires a lot of time overhead.

\end{itemize}

\section{Conclusion}

NER, so as to achieve the purpose of labeling all nested NEs once for all. On the basis of the proposed holistic sentence structure modeling, we further explore NER model enhancement from corpus-aware statistics with the hope of alleviating the serious sparsity issue of NEs. For this purpose, we extend our holistic structure modeling from sentence-level to corpus-level by offering multiple corpus-aware features including Word Freq, Word PMI, POS Freq and POS PMI. The experimental results demonstrate the effectiveness of our proposed model by providing consistent and general performance improvement over strong baselines. In details, our model achieves performance approaching state-of-the-art for two datasets and reaches new state-of-the-art with a large margin for one dataset. Last but not the least, our corpus-level holistic structure modeling shows surprising merit for effective domain adaptation.

\bibliographystyle{acl_natbib}
\bibliography{acl2021}

\end{document}